\documentclass[letterpaper]{article} 
\PassOptionsToPackage{table}{xcolor} 
\usepackage[preprint]{aaai2027}  
\usepackage[hyphens]{url}  
\usepackage{graphicx} 
\usepackage{natbib}  
\usepackage{caption} 
\usepackage{booktabs}
\usepackage{multirow}
\usepackage{booktabs}
\usepackage[most]{tcolorbox}
\tcbuselibrary{breakable}
\usepackage{amssymb}

\usepackage{pifont}
\usepackage{array}

\newcommand{\cmark}{\ding{51}}
\newcommand{\xmark}{\ding{55}}

\tcbset{
    promptbox/.style={
        breakable,
        enhanced,
        boxrule=0.6pt,
        arc=2pt,
        left=4pt,
        right=4pt,
        top=3pt,
        bottom=3pt,
        before skip=4pt,
        after skip=4pt,
        fonttitle=\small\bfseries,
        parbox=false,
        before upper={\parindent0pt},
    }
}

\usepackage{amsmath}
\usepackage{subcaption}
\definecolor{firstgray}{gray}{0.82}
\definecolor{secondgray}{gray}{0.90}
\definecolor{thirdgray}{gray}{0.96}

\newcommand{\first}[1]{%
  \cellcolor{firstgray}%
  \underline{\textbf{#1}}%
  \textsuperscript{\ensuremath{\star}}%
}

\newcommand{\second}[1]{%
  \cellcolor{secondgray}%
  \underline{\textbf{#1}}%
  \textsuperscript{\ensuremath{\dagger}}%
}

\newcommand{\third}[1]{%
  \cellcolor{thirdgray}%
  \underline{\textbf{#1}}%
  \textsuperscript{\ensuremath{\ddagger}}%
}

\newcommand{\modelcite}[2]{%
  \shortstack[l]{#1\\[-1pt]\citet{#2}}%
}

\title{Behavior2Value: Benchmarking and Empowering LLMs for Consumer Value Measurement from E-commerce Behaviors}

\author{
  Peixuan Hou\textsuperscript{\rm 1},
  Bin Chen\textsuperscript{\rm 1},
  Li He\textsuperscript{\rm 2},
  Jian Xu\textsuperscript{\rm 2},
  Bo Zheng\textsuperscript{\rm 2},
  Xiuli Ma\textsuperscript{\rm 1},
  Guojie Song\textsuperscript{\rm 1}\corresponding
}

\affiliations{
  \textsuperscript{\rm 1}State Key Laboratory of General Artificial Intelligence,\\
  School of Intelligence Science and Technology, Peking University\\
  \textsuperscript{\rm 2}Alibaba Group\\
  gjsong@pku.edu.cn
}

\begin{document}
\maketitle

\begin{abstract}

Human values are deep motivational orientations that shape human behaviors. In e-commerce, they reveal the stable drivers behind users' purchase decisions. Compared with short-term interests, consumer values better explain how users evaluate products before purchase. However, consumer values are often implicit in complex and fragmented behavioral trajectories, leaving value measurement from e-commerce behaviors largely underexplored. To this end, we propose the \textbf{Behavior-to-Value (B2V)} task, which aims to identify consumer values from e-commerce behavioral trajectories. Centered on this task, we first construct the \textbf{E-commerce Consumption Value Taxonomy (ECVT)} and introduce \textbf{B2V-Bench}, the first B2V dataset and benchmark, based on anonymized Taobao behavioral logs. B2V-Bench consists of real-world purchase decision episodes, covering 25 types of purchase behaviors, along with corresponding consumer value orientations manifested in each episode. To improve consumer value measurement accuracy, we further present \textbf{B2V-Verifier}, a behavior-to-value measurement model based on Value Verification Tuning, which learns to assess whether behaviors provide sufficient evidence for each value inference. Experiments show that B2V-Verifier outperforms strong LLM baselines, improving multi-label classification by 34\%. The dataset and code will be publicly released upon acceptance.

\end{abstract}

\begin{figure*}[t!]
    \centering    
    \includegraphics[width=\linewidth, keepaspectratio]{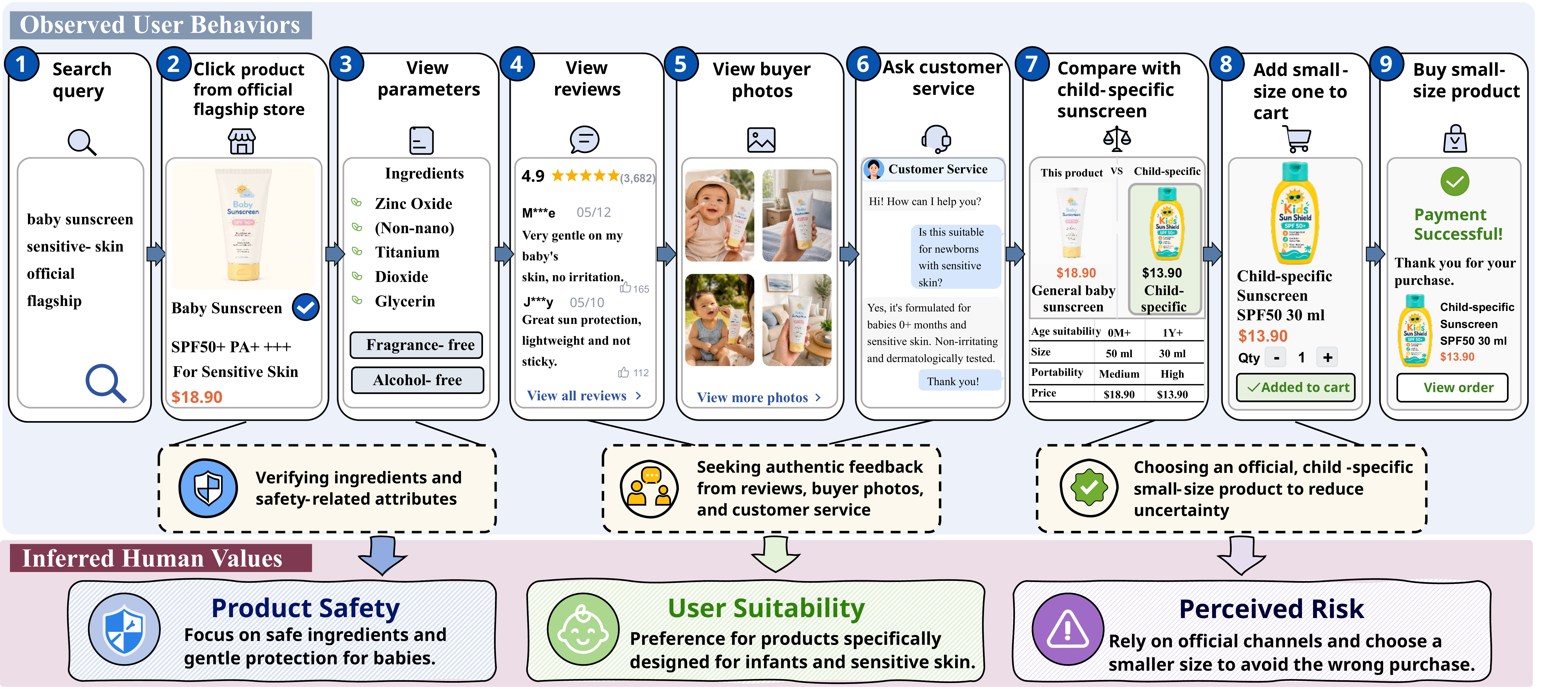}
    \caption{Illustration of the proposed \textbf{Behavior-to-Value} task. B2V measures the consumer value tendencies reflected in a specific episode of e-commerce behavior, enabling more interpretable consumer behavior analysis and personalized services.
    }
    \label{fig:b2v_task}
    \vspace{-1em}
\end{figure*}

\section{Introduction}

Human values are stable beliefs about desirable goals and principles~\cite{rokeach1973nature, schwartz1992universals}. Unlike short-term interests, they reflect deeper motivational orientations that shape how people interpret situations, compare alternatives, and ultimately take action. Therefore, studying values provides a foundation for explaining stable and transferable patterns in human behavior~\cite{bardi2003values}.


Values broadly influence human life and play a profound role in consumer behavior~\cite{manolios2019influence,ge2020learning}. In e-commerce, for example, risk-averse users may repeatedly inspect reviews and compare buyer feedback before purchase to reduce decision uncertainty, while brand-oriented users may prefer official stores and well-known brands. These value-driven behavioral patterns are crucial for e-commerce, as they reveal users' enduring priorities, account for behavioral consistency across situations, and support more accurate and personalized e-commerce services~\cite{zhang2020explainable}. Nevertheless, consumer values are often implicit in complex and fragmented behavioral traces, leaving value measurement from e-commerce behaviors largely underexplored. To fill this gap, we propose the \textbf{Behavior-to-Value (B2V)} task, which aims to identify the consumer value tendencies manifested in an episode of e-commerce behavior, as illustrated in Figure~\ref{fig:b2v_task}. This task provides deeper support for user understanding, consumption behavior analysis and personalized services.

However, existing methods are limited for the B2V task. Traditional sequential models~\cite{hidasi2015session,zhou2018deep,sun2019bert4rec} struggle to capture value orientations embedded in product attributes and behavioral actions. Large language models (LLMs)~\cite{comanici2025gemini,gpt5,liu2025deepseek,yang2025qwen3} provide stronger semantic reasoning capabilities~\cite{zhao2023survey}, yet without task-specific data and explicit evidence constraints, they may still rely on superficial behavior--value associations and produce unreliable value judgments. Overall, B2V faces two key challenges: \textbf{(1) Data scarcity.} Existing e-commerce datasets~\cite{jin2024shopping,ben2015yoochoose,jin2023amazon} lack consumer value annotations, while no e-commerce value taxonomy exists. Moreover, they cover only a few coarse-grained behavior types and treat isolated interactions as individual samples, obscuring users' pre-purchase deliberation. \textbf{(2) Value identification difficulty.} Values are latent and cannot be inferred through a simple ``behavior-to-value'' mapping. Reliable value identification requires assessing whether behavioral evidence sufficiently supports a value orientation; otherwise, models may over-attribute values from isolated cues while ignoring counter-evidence.

To tackle these challenges, we develop the \textbf {E-commerce Consumption Value Taxonomy (ECVT)}, comprising 30 fine-grained value constructs tailored to e-commerce. It is grounded in real-world e-commerce reviews, refined with expert knowledge from established psychological value scales. Based on ECVT, we construct \textbf{B2V-Bench}, the first dataset and benchmark for the B2V task. B2V-Bench contains 5,440 real-world purchase-decision episodes derived from anonymized behavioral logs on Taobao, one of the world's largest e-commerce platforms. Centered on a completed purchase, each episode contains the multi-step pre-purchase behaviors leading to that purchase, covering 25 behavior types such as queries, clicks, review inspection, detail-page browsing, favorites, and add-to-cart actions. For each episode, B2V-Bench provides a summary of the behavior sequence, value labels from our ECVT, and annotation rationales.

Building on this dataset, we propose \textbf{B2V-Verifier}, a behavior-to-value measurement model based on Value Verification Tuning. Rather than directly mapping behavioral trajectories to values, B2V-Verifier learns to verify whether a value label is sufficiently supported by behavioral evidence, thereby producing more accurate and reliable value predictions. Extensive experiments show that B2V-Verifier consistently outperforms multiple advanced LLMs, with approximate gains of 34\% in multi-label classification, 7.4\% in label ranking, and 5.5\% in primary-label identification.


In summary, the key contributions are as follows:
\begin{itemize}
    \item We introduce the \textbf{Behavior-to-Value} task, which identifies consumer value orientations from e-commerce behavior trajectories and provides a new perspective for consumption behavior analysis and personalized services.
    \item The \textbf{E-commerce Consumption Value Taxonomy} is proposed to provide a label space for consumer values in e-commerce. Based on ECVT, we construct \textbf{B2V-Bench}, a dataset and benchmark comprising real-world e-commerce behavioral trajectories, each paired with its summary, value labels, and annotation rationales.
    \item We present \textbf{B2V-Verifier} as a behavior-to-value measurement model based on Value Verification Tuning, learning to assess whether behaviors provide sufficient evidence for each value inference. Experiments show that B2V-Verifier outperforms strong baselines, with improvements of 34\% in multi-label classification.

\end{itemize}

\begin{table*}[t]
    \centering
    \scriptsize
    \setlength{\tabcolsep}{2.4pt}
    \renewcommand{\arraystretch}{0.9}
    \begin{tabular}{@{}c p{0.285\textwidth} p{0.135\textwidth} c p{0.190\textwidth} c >{\centering\arraybackslash}p{0.105\textwidth} p{0.060\textwidth}@{}}
        \toprule
        \textbf{Type}
        & \textbf{Dataset}
        & \textbf{Task}
        & \textbf{Behavior Type}
        & \textbf{Annotation Unit}
        & \textbf{R}
        & \textbf{Value Labels}
        & \textbf{Source} \\
        \midrule

        \multirow{9}{*}{\rotatebox{90}{\textit{E-commerce}}}
        & Amazon Review~\citep{keung2020multilingualamazonreviewscorpus}
        & Review prediction
        & $1$
        & per-item ($1$ step)
        & \xmark
        & \xmark
        & Amazon \\

        & Amazon-M2~\citep{jin2023amazon}
        & Recommendation
        & $1$
        & per-session (avg. $4.2$ steps)
        & \xmark
        & \xmark
        & Amazon \\

        & SessionIntentBench~\citep{yang2026sessionintentbench}
        & Intention-shift modeling
        & $1$
        & per-task (avg. $3.4$ steps)
        & \xmark
        & \xmark
        & Amazon \\

        & Repeat Buyers~\citep{liu2016repeatbuyers}
        & Buyer prediction
        & $4$
        & per-click ($1$ step)
        & \xmark
        & \xmark
        & Tmall \\

        & Taobao~\citep{Zhu_2018}
        & Recommendation
        & $4$
        & per-click ($1$ step)
        & \xmark
        & \xmark
        & Taobao \\

        & YOOCHOOSE~\citep{ben2015yoochoose}
        & Purchase prediction
        & $2$
        & per-session (avg. $3.5$ steps)
        & \xmark
        & \xmark
        & Retailer \\

        & Shopping~MMLU~\citep{jin2024shopping}
        & Recommendation
        & $3$
        & per-item ($1$ step)
        & \xmark
        & \xmark
        & Amazon \\

        & MerRec~\citep{li2024merreclargescalemultipurposemercari}
        & Multi-task
        & $5$
        & per-session (avg. $5.6$ steps)
        & \xmark
        & \xmark
        & Mercari \\

        & OPeRA~\citep{wang2026operadatasetobservationpersona}
        & Behavior simulation
        & $8$
        & per-action ($1$ step)
        & \xmark
        & \xmark
        & Study \\

        \midrule

        \multirow{2}{*}{\rotatebox{90}{\textit{Value}}}
        & ValueBench~\citep{ren2024valuebench}
        & Value probing (LLM)
        & \xmark
        & per-item ($1$ step)
        & \xmark
        & \textit{10, Schwartz}
        & Synthetic \\

        & PVQ~\citep{schwartz2001extending}
        & Self-report PVQ
        & \xmark
        & per-questionnaire
        & \xmark
        & \textit{10, Schwartz}
        & Survey \\

        \midrule

        \rowcolor{gray!12}
        \textbf{Ours}
        & \textbf{B2V-Bench}
        & \textbf{B2V}
        & $\mathbf{25}$
        & \textbf{per-session (avg. $341$ steps)}
        & \cmark
        & \textbf{28, ECVT}
        & Taobao \\

        \bottomrule
    \end{tabular}
    \caption{
        Comparison of existing e-commerce and value datasets with B2V-Bench. \textbf{Annotation Unit}: the data instance being annotated with the average number of behavior steps per unit. \textbf{R}: whether annotations include explicit supporting reasons.
    }
    \label{tab:dataset_comparison}
    \vspace{-1em}
\end{table*}

\vspace{-0.2em}
\section{B2V-Bench: A New Dataset for B2V}

Although existing e-commerce datasets are widely used for recommendation and behavior modeling, they are not well suited for the B2V task. As summarized in Table~\ref{tab:dataset_comparison}, they have three limitations: (1) no value-level labels, (2) limited behavior types, and (3) individual interactions as annotation units. These constraints make them insufficient for inferring consumption values from users' decision processes.

To address this gap, we propose the \textbf {E-commerce Consumption Value Taxonomy (ECVT)}, comprising 30 fine-grained value constructs tailored to e-commerce. Based on ECVT, we introduce \textbf{B2V-Bench}, which contains 5,440 samples. Each sample is built around a real purchase-decision episode, which consists of multi-step pre-purchase behaviors leading to the same completed purchase. The sample further includes a summary of the behavioral trajectory, value labels, and rationale evidence supporting these labels. B2V-Bench captures richer behavioral evidence with 25 types of pre-purchase actions, including search, click, add-to-cart, coupon inspection, specification checking, review browsing, Q\&A viewing and customer service interaction (see details in Appendix C). We annotate behavior sequences that capture a complete purchase decision process (341 actions on average), because values emerge from the overall process.




\vspace{-0.5em}
\subsection{E-commerce Consumption Value Taxonomy}

\begin{figure}[t]
    \centering    
    \includegraphics[width=0.65\linewidth, keepaspectratio]{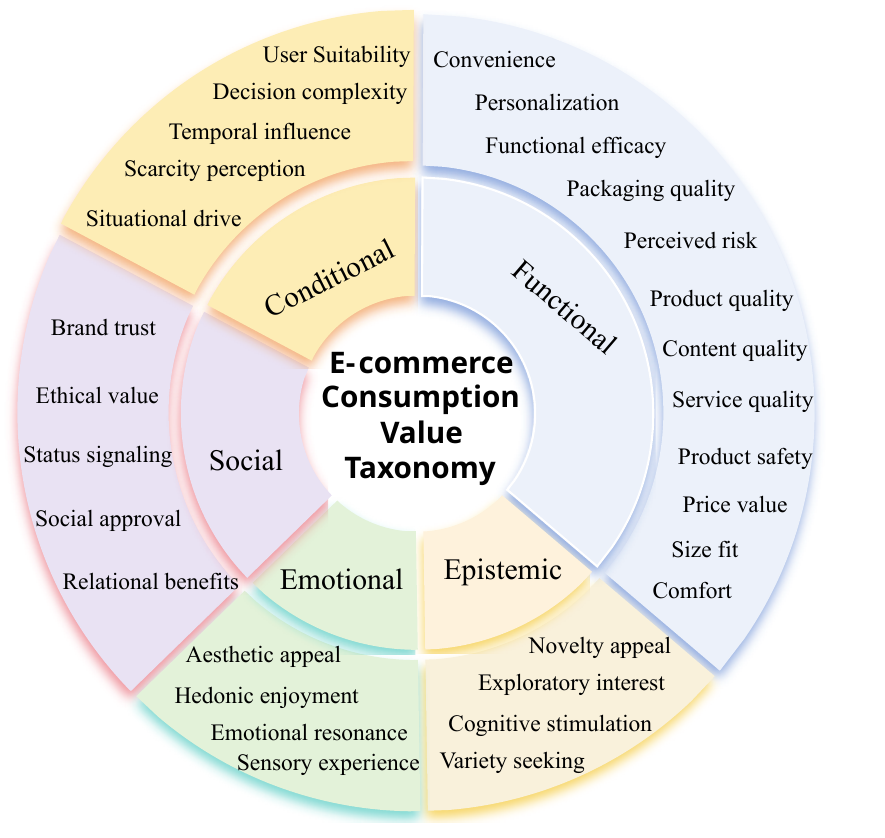}
    \caption{Overview of ECVT, comprising 30 fine-grained consumer value constructs across the five TCV dimensions.}
      \label{fig:ECVT}
\vspace{-1.5em}
\end{figure}

Constructing a B2V dataset requires an e-commerce-specific value taxonomy. Classic consumption value theories~\cite{sweeney2001consumer,holbrook1999consumer,sheth1991we} are too coarse-grained to capture the fine-grained motivations in modern online commerce. For example, the Theory of Consumption Values (TCV)~\cite{sheth1991we} may broadly treat review reading and coupon checking as functional value, while in e-commerce they reflect distinct concerns such as risk aversion and price sensitivity. E-commerce reviews provide a natural basis for taxonomy construction, as they capture users' post-purchase evaluations in real consumption contexts. Accordingly, we construct ECVT by deriving value dimensions from large-scale reviews and refining them with established consumption value scales. As shown in Figure~\ref{fig:ECVT}, ECVT comprises 30 fine-grained value constructs organized into the five dimensions of the Theory of Consumption Values (TCV).



For bottom-up value discovery, we use Amazon Reviews 2023 as empirical grounding to derive candidate value dimensions from real consumption contexts. We split each review into independent statements that express specific evaluations. Leveraging the semantic understanding and world knowledge of LLMs, we organize these statements into candidate value dimensions and describe each by its meaning, decision role, polarity, and broader value category. To provide theoretical grounding, we consolidate established consumption value scales from major frameworks~\cite{sheth1991we,sweeney2001consumer,holbrook1999consumer,t1_blut2024customer,t2_fornell1996american,t3_parasuraman2005multiple,t4_loiacono2007webqual,t5_reimers2019sentence,t6_zheng2023judging}. We then cluster semantically similar candidates and invite psychology experts to refine them with reference to these scales, removing redundancy and noise while preserving meaningful long-tail values, ultimately yielding 30 value constructs.

We also conduct a comprehensive psychometric validation of ECVT.
The results in Appendix D show that ECVT is semantically coherent, clearly differentiated, broadly applicable across product categories, and theoretically grounded, while capturing fine-grained values specific to e-commerce.

\begin{figure*}[t]
    \centering    
    \includegraphics[width=\linewidth, keepaspectratio]{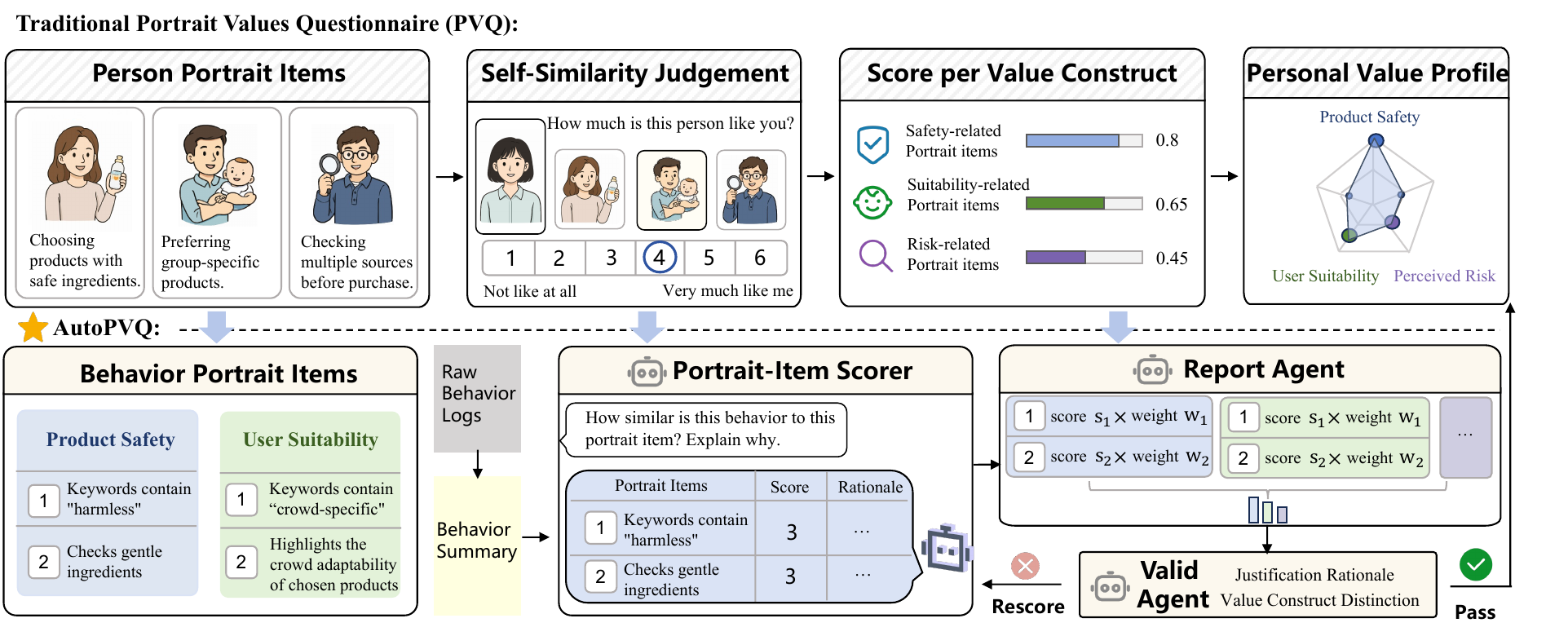}
    \caption{Framework of AutoPVQ. AutoPVQ adapts PVQ-style value measurement to e-commerce behavior annotation through behavior-item scoring, construct-level aggregation, and evidence-based label verification.}
      \label{fig:autopvq}
\vspace{-1em}
\end{figure*}

\vspace{-0.5em}
\subsection{Behavior-to-Value Annotation}

We obtained anonymized user behavior data from Taobao. Each behavior episode consists of a completed order and the user's same-category interactions during the preceding day, capturing a coherent purchase goal and serving as the basic unit for value annotation. The key challenge is to infer latent consumption values from behaviors. Traditional self-report value questionnaires~\citep{rokeach1973nature,schwartz1992universals,schwartz2001extending} provide a mature measurement paradigm but are costly to administer and unsuitable for large-scale behavioral data. We therefore design a four-stage annotation process: (1) behavior summarization condenses redundant raw logs into concise summaries; (2) a protocol is developed to define annotation rules and behavioral portrait items that describe the behaviors supporting each ECVT construct; (3) AutoPVQ, a multi-agent framework that adapts the measurement logic of the Portrait Values Questionnaire (PVQ)~\citep{schwartz2001extending}, matches the summaries against these behavioral portrait items to generate initial annotations; and (4) human experts verify and refine the results.

\noindent\textbf{Behavior Summarization.} Direct annotation over raw behavior logs is costly and error-prone, as many interactions are redundant and obscure key evidence underlying the purchase decision. We therefore compress each episode into a standardized behavioral summary that preserves critical decision signals, including product comparison, information verification, candidate switching, and final choice. To preserve the purchase-decision process, we compare multiple LLMs and prompt variants through blinded expert evaluation. GPT-5.4 achieves the highest overall quality and is therefore used to generate the behavioral representations for annotation. Full evaluation details are provided in Appendix G.

\noindent\textbf{Value Construct Operationalization Protocol.} To provide AutoPVQ with behavioral portrait items and clear annotation rules, we develop the Value Construct Operationalization Protocol (VCOP). VCOP specifies the behaviors that support each value and the criteria for distinguishing similar values, enabling reliable and theoretically grounded annotation.


For each value construct in ECVT, VCOP describes its core meaning, typical behavioral evidence, referred to as behavioral portrait items, possible counter-evidence, and distinctions from similar constructs. It is developed through a theory-driven and empirically refined process: initial guidelines are derived from established value scales, e-commerce review data, and consumer psychology literature, and then iteratively improved through pilot annotation and disagreement analysis. This process supports standardized and reliable annotation (see details in Appendix F).



\begin{table}[t]
\fontsize{14pt}{18pt}\selectfont
\centering
\captionsetup{skip=5pt}
\resizebox{\linewidth}{!}{
\begin{tabular}{lccc}
\toprule[0.12em]
Metric 
& GPT-5.4 
& Claude-Sonnet 
& Qwen-3.5 \\
\midrule

Primary Agreement (\%) $\uparrow$
& \textbf{86.7}
& 82.9
& 79.6 \\

Label-set Jaccard $\uparrow$
& \textbf{0.812}
& 0.774
& 0.741 \\

Exact Match (\%) $\uparrow$
& \textbf{72.4}
& 67.1
& 63.5 \\

Cohen's $\kappa$ $\uparrow$
& \textbf{0.781}
& 0.735
& 0.692 \\

\bottomrule[0.12em]
\end{tabular}
}
\caption{Cross-model and human validation results of AutoPVQ on 1,000 behavior episodes.}
\label{tab:autopvq_validation}
\vspace{-1em}
\end{table}

\noindent\textbf{AutoPVQ Annotation.} PVQ measures values by assessing how closely a person matches portraits that describe value-related priorities. Following this logic, AutoPVQ compares each behavioral summary with the behavioral portrait items of each value construct. As shown in Figure~\ref{fig:autopvq}, it decomposes annotation into specialized steps: portrait-item scoring, discriminative verification, feedback refinement, and result aggregation. Cross-agent verification and iterative refinement improve the reliability of the initial annotations.


Based on the behavior summary, the Portrait-Item Scorer first evaluates how well the episode matches the behavioral portrait items defined in VCOP. Because the value implications of e-commerce behaviors often depend on product attributes and category semantics, the scorer leverages the semantic knowledge of LLMs to contextualize observable evidence. For example, cues such as ``portable'' and ``installation-free'' may support the construct of convenience. Such reasoning is used only to interpret evidence, while the final judgment remains constrained by VCOP.

The Report Agent combines the support scores of individual behavioral portrait items into an overall score for each value construct. For each value construct \(v\) with associated VCOP items \(I_v\), the raw score is computed as:
\[
\mathrm{Raw}(v)=\frac{\sum_{i \in I_v} w_i s_i}{\sum_{i \in I_v} w_i},
\]
where \(s_i \in \{0,1,2,3\}\) is the support score for each item and \(w_i\) is its diagnostic weight, reflecting evidence strength. To measure the relative salience of each value within an episode, we apply episode-level centering:
\[
\mathrm{Score}(v)
=
\mathrm{Raw}(v)
-
\frac{1}{|V|}
\sum_{v' \in V}
\mathrm{Raw}(v'),
\]
where \(V\) is the candidate value set. The top construct is assigned as the primary label if its score exceeds \(\tau_1\); secondary labels are assigned when additional constructs exceed \(\tau_2\) and have independent evidence. The final output includes value labels, a quality score, and an evidence summary. To reduce over-inference and construct confusion, the Valid Agent verifies each candidate label against VCOP distinction rules and counter-evidence. Insufficiently supported labels are rescored by the Portrait-Item Scorer.

\begin{figure*}[t]
\centering
\includegraphics[width=\linewidth]{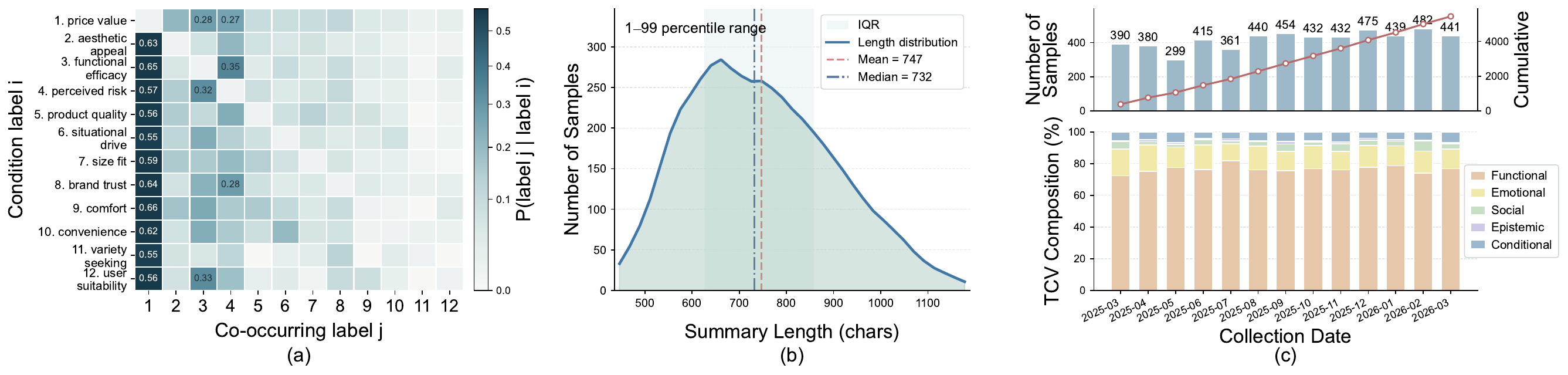}
\caption{Characteristics of B2V-Bench. 
(a) Label co-occurrence patterns among frequent value constructs. 
(b) Distribution of behavioral summary lengths. 
(c) Monthly data volume and TCV composition over time.}
\label{fig:character}
\vspace{-0.5em}
\end{figure*}

\begin{figure*}[t]
    \centering    
    \includegraphics[width=\linewidth, keepaspectratio]{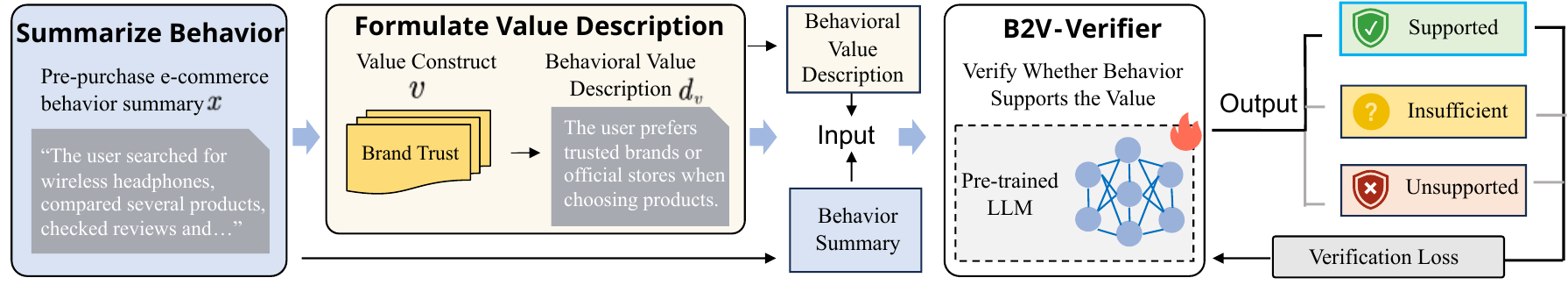}
    \caption{Overview of Value Verification Tuning. Given a behavior summary and verbalized value descriptions, B2V-Verifier determines whether the observed behavior supports the value and predicts a three-way verification label.}
      \label{fig:vvf}
\vspace{-1em}
\end{figure*}

\noindent\textbf{Expert Review and Correction.} We randomly sample 1,000 behavior episodes as an expert-annotated reference set and invite three psychology experts to jointly annotate them from scratch and resolve disagreements through discussion to reach consensus. To select the backbone for AutoPVQ, we instantiate three advanced LLMs under the same multi-agent protocol and compare their outputs with expert annotations. As shown in Table~\ref{tab:autopvq_validation}, GPT-5.4 achieves the best overall alignment with expert judgments and is adopted as the backbone for full-scale AutoPVQ annotation. Human annotators further review and revise its outputs following VCOP guidelines. They assess label validity, evidence sufficiency, and potential over-inference, and correct unsupported or ambiguous value labels to produce the final human-verified dataset.

\vspace{-0.5em}
\subsection{Dataset Evaluation and Analysis}

\noindent\textbf{Dataset Split and Quality Control.} B2V-Bench is collected from 13 monthly snapshots spanning March 2025 to March 2026, comprising 5,440 annotated episodes. A total of 4,352 episodes are used for training and 544 for validation, with labels generated by AutoPVQ and then reviewed and corrected by human annotators. The 544-episode test set is sampled from a pool of 1,000 episodes annotated from scratch by psychology experts, ensuring model-independent evaluation.



\noindent\textbf{Dataset Characteristics.} B2V-Bench covers a broad and fine-grained label space of consumer values. It uses 28 value labels from ECVT, with an additional \emph{none} tag for evidence-insufficient episodes to avoid forced value attribution. After expert discussion, \emph{decision complexity} and \emph{exploratory interest} are excluded from the default label set because they are difficult to identify reliably from a single behavior episode. B2V-Bench exhibits a clear multi-label structure: 75.2\% of episodes include at least one secondary value, indicating that purchase decisions often reflect multiple value orientations. Figure~\ref{fig:character}(a) shows clear asymmetric co-occurrence patterns among frequent value labels, such as \emph{functional efficacy} with \emph{perceived risk} and \emph{brand trust} with \emph{perceived risk}. These directional dependencies suggest that consumer values often appear in primary--secondary relations, supporting our multi-label formulation and the need for disambiguation rules. We further examine the evidence sufficiency and temporal stability of B2V-Bench. Figure~\ref{fig:character}(b) shows that most behavioral summaries contain 500--1000 characters, providing sufficient context for value inference. Figure~\ref{fig:character}(c) shows comparable session volumes and stable TCV composition across monthly snapshots, suggesting that B2V-Bench captures consistent patterns rather than seasonal artifacts.

\begin{table*}[!t]
\centering
\scriptsize
\setlength{\tabcolsep}{2.3pt}
\renewcommand{\arraystretch}{0.8}
\resizebox{\textwidth}{!}{
\begin{tabular}{lc|cccccc|ccc|ccc}
\toprule
\textbf{Model} & \textbf{Prompt}
& \multicolumn{6}{c|}{\textbf{Multi-label Classification}}
& \multicolumn{3}{c|}{\textbf{Label Ranking}}
& \multicolumn{3}{c}{\textbf{Label Identification}} \\
\cmidrule(lr){3-8} \cmidrule(lr){9-11} \cmidrule(lr){12-14}
& & \textbf{Ma-F1} \(\uparrow\) 
& \textbf{Mi-F1} \(\uparrow\) 
& \textbf{W-F1} \(\uparrow\) 
& \textbf{Prec.} \(\uparrow\) 
& \textbf{Rec.} \(\uparrow\) 
& \textbf{Ham.} \(\downarrow\) 
& \textbf{N@1} \(\uparrow\) 
& \textbf{N@2} \(\uparrow\) 
& \textbf{N@3} \(\uparrow\) 
& \textbf{Acc.} \(\uparrow\) 
& \textbf{T2 Acc.} \(\uparrow\)
& \textbf{Avg.} \(\uparrow\) \\
\midrule

\multirow{4}{*}{\modelcite{Claude-Haiku-4.5}{anthropic2025haiku45}}
& P1 & 0.256 & 0.510 & 0.559 & 0.313 & 0.407 & 0.083 & 0.671 & 0.610 & 0.633 & 0.548 & 0.677 & 0.613 \\
& P2 & 0.286 & 0.576 & 0.600 & 0.315 & 0.393 & 0.068 & 0.714 & 0.672 & 0.672 & 0.574 & 0.743 & 0.659 \\
& P3 & 0.313 & 0.596 & 0.612 & 0.367 & 0.398 & 0.060 & 0.730 & 0.692 & 0.666 & 0.593 & 0.770 & 0.682 \\
& P4 & 0.275 & 0.624 & 0.622 & 0.301 & 0.355 & \third{0.053} & 0.757 & 0.727 & 0.679 & 0.604 & \third{0.777} & 0.691 \\
\cmidrule(lr){1-14}

\multirow{4}{*}{\modelcite{DeepSeek-V3.2}{liu2025deepseek}}
& P1 & 0.266 & 0.568 & 0.605 & 0.300 & 0.433 & 0.073 & 0.717 & 0.676 & 0.688 & 0.569 & 0.732 & 0.651 \\
& P2 & 0.289 & 0.597 & 0.611 & 0.267 & 0.456 & 0.065 & 0.721 & 0.697 & 0.692 & 0.561 & 0.757 & 0.659 \\
& P3 & 0.317 & 0.619 & 0.645 & 0.318 & 0.442 & 0.059 & 0.757 & 0.711 & 0.701 & 0.595 & 0.757 & 0.676 \\
& P4 & 0.350 & \second{0.655} & 0.659 & 0.380 & 0.422 & \second{0.050} & 0.773 & \second{0.749} & 0.711 & 0.613 & \second{0.781} & \second{0.697} \\
\cmidrule(lr){1-14}

\multirow{4}{*}{\modelcite{Gemini-2.5}{comanici2025gemini}}
& P1 & 0.276 & 0.582 & 0.570 & 0.350 & 0.353 & 0.067 & 0.773 & 0.691 & 0.691 & 0.563 & 0.695 & 0.629 \\
& P2 & 0.248 & 0.583 & 0.549 & 0.369 & 0.339 & 0.064 & 0.775 & 0.703 & 0.675 & 0.578 & 0.717 & 0.648 \\
& P3 & 0.263 & 0.596 & 0.568 & 0.350 & 0.355 & 0.060 & \second{0.781} & 0.713 & 0.674 & 0.612 & 0.734 & 0.673 \\
& P4 & 0.242 & 0.604 & 0.572 & 0.305 & 0.301 & 0.054 & 0.775 & 0.726 & 0.657 & 0.615 & 0.751 & 0.683 \\
\cmidrule(lr){1-14}

\multirow{4}{*}{\modelcite{GLM-5}{zeng2026glm5}}
& P1 & 0.308 & 0.585 & 0.612 & 0.297 & 0.433 & 0.066 & 0.708 & 0.679 & 0.676 & 0.569 & 0.704 & 0.637 \\
& P2 & 0.335 & 0.609 & 0.628 & 0.354 & 0.415 & 0.058 & 0.704 & 0.673 & 0.659 & 0.578 & 0.714 & 0.646 \\
& P3 & 0.339 & 0.625 & 0.637 & 0.379 & 0.389 & \third{0.053} & 0.673 & 0.666 & 0.647 & 0.556 & 0.717 & 0.637 \\
& P4 & 0.358 & 0.608 & 0.622 & \second{0.417} & 0.405 & 0.055 & 0.665 & 0.647 & 0.630 & 0.558 & 0.703 & 0.631 \\
\cmidrule(lr){1-14}

\multirow{4}{*}{\modelcite{GPT-4o-mini}{openai2024gpt4omini}}
& P1 & 0.248 & 0.514 & 0.525 & 0.297 & 0.316 & 0.083 & 0.710 & 0.642 & 0.641 & 0.517 & 0.638 & 0.578 \\
& P2 & 0.253 & 0.504 & 0.521 & 0.303 & 0.298 & 0.084 & 0.721 & 0.642 & 0.635 & 0.526 & 0.645 & 0.586 \\
& P3 & 0.248 & 0.503 & 0.519 & 0.350 & 0.298 & 0.085 & 0.717 & 0.644 & 0.635 & 0.543 & 0.658 & 0.601 \\
& P4 & 0.220 & 0.569 & 0.577 & 0.305 & 0.273 & 0.069 & 0.691 & 0.657 & 0.659 & 0.532 & 0.704 & 0.618 \\
\cmidrule(lr){1-14}

\multirow{4}{*}{\modelcite{GPT-5}{gpt5}}
& P1 & 0.314 & 0.556 & 0.613 & 0.284 & 0.447 & 0.075 & 0.599 & 0.623 & 0.651 & 0.476 & 0.701 & 0.589 \\
& P2 & \third{0.400} & 0.597 & 0.654 & 0.362 & \second{0.531} & 0.067 & 0.686 & 0.687 & 0.705 & 0.548 & 0.749 & 0.649 \\
& P3 & 0.360 & \third{0.629} & \second{0.673} & 0.342 & 0.467 & 0.062 & 0.690 & 0.697 & 0.725 & 0.548 & 0.775 & 0.662 \\
& P4 & \second{0.417} & 0.628 & \third{0.666} & \third{0.400} & \third{0.524} & 0.060 & 0.703 & 0.704 & 0.719 & 0.558 & 0.770 & 0.664 \\
\cmidrule(lr){1-14}

\multirow{4}{*}{\modelcite{Kimi-K2.6}{moonshot2026k2_6}}
& P1 & 0.275 & 0.578 & 0.613 & 0.273 & 0.391 & 0.072 & 0.762 & 0.726 & 0.717 & \third{0.621} & 0.760 & 0.691 \\
& P2 & 0.325 & 0.585 & 0.598 & 0.360 & 0.432 & 0.070 & \third{0.779} & 0.729 & 0.723 & \second{0.632} & 0.758 & \third{0.695} \\
& P3 & 0.307 & 0.593 & 0.620 & 0.374 & 0.398 & 0.069 & 0.757 & 0.724 & \third{0.726} & 0.604 & 0.764 & 0.684 \\
& P4 & 0.304 & 0.591 & 0.615 & 0.338 & 0.381 & 0.066 & 0.716 & 0.704 & 0.695 & 0.568 & 0.737 & 0.653 \\
\cmidrule(lr){1-14}

\multirow{4}{*}{\modelcite{Qwen3-235B}{yang2025qwen3}}
& P1 & 0.253 & 0.559 & 0.577 & 0.233 & 0.404 & 0.075 & 0.757 & 0.691 & 0.696 & 0.599 & 0.727 & 0.663 \\
& P2 & 0.271 & 0.565 & 0.582 & 0.280 & 0.386 & 0.074 & 0.730 & 0.700 & 0.698 & 0.574 & 0.742 & 0.658 \\
& P3 & 0.290 & 0.581 & 0.602 & 0.300 & 0.410 & 0.071 & 0.747 & 0.715 & 0.713 & 0.595 & 0.738 & 0.667 \\
& P4 & 0.288 & 0.605 & 0.622 & 0.310 & 0.399 & 0.066 & 0.757 & \third{0.731} & \second{0.732} & 0.593 & 0.764 & 0.679 \\

\midrule
\textbf{B2V-Verifier (Ours)} & P3
& \first{0.615}
& \first{0.815}
& \first{0.812}
& \first{0.659}
& \first{0.591}
& \first{0.028}
& \first{0.824}
& \first{0.807}
& \first{0.798}
& \first{0.652}
& \first{0.842}
& \first{0.747} \\
\bottomrule
\end{tabular}
}
\caption{Performance comparison of mainstream LLM baselines and B2V-Verifier on B2V-Bench. \(P1\)--\(P4\) denote label-only, +definition, +behavioral value description, and +few-shot prompting settings, respectively. The top three results are highlighted with dark, medium, and light gray backgrounds, respectively, and are marked with \(\star\), \(\dagger\), and \(\ddagger\). For B2V-Verifier,  \(P3\) corresponds to its standard input format used in VVT so we evaluate it under the same setting as training. More results are in Appendix B.}
\label{tab:b2v_results}
\vspace{-1em}
\end{table*}

\begin{table*}[t] 
\centering
\small
\resizebox{\textwidth}{!}{
\begin{tabular}{l|ccc|cc|cc}
\toprule
\textbf{Model Variant} 
& \textbf{Macro-F1} 
& \textbf{Micro-F1} 
& \textbf{Weighted-F1} 
& \textbf{NDCG@1} 
& \textbf{NDCG@3} 
& \textbf{Primary Acc.} 
& \textbf{Top-2 Acc.} \\
\midrule
B2V-Verifier (Full VVT) 
& \textbf{0.615} 
& \textbf{0.815} 
& \textbf{0.812} 
& \textbf{0.824} 
& \textbf{0.798} 
& \textbf{0.652} 
& \textbf{0.842} \\

w/o Verification Formulation 
& 0.548 
& 0.766 
& 0.759 
& 0.773 
& 0.742 
& 0.589 
& 0.796 \\

w/o Value Description 
& 0.574 
& 0.789 
& 0.784 
& 0.794 
& 0.764 
& 0.615 
& 0.813 \\

w/o Insufficient Modeling 
& 0.563 
& 0.781 
& 0.776 
& 0.785 
& 0.755 
& 0.603 
& 0.807 \\

w/o Unsupported Modeling
& 0.582 
& 0.795 
& 0.789 
& 0.802 
& 0.772 
& 0.624 
& 0.819 \\

\bottomrule
\end{tabular}
}
\caption{Ablation study of B2V-Verifier. The full VVT framework consistently outperforms all ablations.
}
\label{tab:ablation}
\vspace{-1em}
\end{table*}

\vspace{-0.3em}
\section{Method: B2V-Verifier}

Building on B2V-Bench, we develop \textbf{B2V-Verifier} by applying \textbf{Value Verification Tuning (VVT)} to Qwen3-8B. Unlike standard label classification, measuring consumption values from behavior requires context-dependent inference from indirect behavioral evidence, since identical surface behaviors may reflect different motivations. For example, a high-priced purchase does not necessarily imply low price sensitivity, nor does checking reviews always indicate risk aversion. So, B2V prediction should not only associate behaviors with value labels, but also determine whether the observed trajectory provides sufficient evidence for a specific value inference.


Motivated by this observation, we design Value Verification Tuning (VVT), which reformulates B2V prediction as a value-wise verification task. As illustrated in Figure~\ref{fig:vvf}, for each value construct, we derive a \textit{behavioral value description} from VCOP that defines the construct and specifies the concrete behaviors that support it. Given a behavior summary and a candidate value description, the model assesses whether the observed trajectory provides sufficient evidence for that value. This formulation decomposes multi-label prediction into explicit evidence judgments for individual value constructs, rather than directly generating a likely label set. It therefore preserves the multi-label nature of B2V while requiring each predicted value to be independently justified by the behavior, reducing unsupported attribution.

Formally, given a behavior summary $x$ and a value construct $v$, we derive a behavioral value description $d_v$ that defines $v$ and specifies the concrete behaviors that support it. The model takes $(x,d_v)$ as input and predicts a verification label $y$, indicating how well the observed behavior supports the candidate value:
\[
y \in \{\textit{supported}, \textit{unsupported}, \textit{insufficient}\}.
\]
Specifically, \textit{supported} is assigned when the trajectory contains clear evidence for the candidate value. A prediction is considered \textit{unsupported} if the value is irrelevant to, contradicted by, or inconsistent with the observed behavior. When the value appears plausible but the available evidence remains too weak or incomplete to justify the inference, the model outputs \textit{insufficient}.
This distinction is important because a behavior may appear consistent with a value without providing enough evidence to confirm it. Explicitly modeling insufficient evidence reduces uncertain or unsupported inferences, enabling more accurate value measurement.


We construct VVT training instances from B2V-Bench by pairing each behavior trajectory with descriptions of candidate values. Annotated value labels form \textit{supported} instances. Labels proposed but rejected by the validation agent are assigned \textit{insufficient}, as they represent plausible interpretations without adequate behavioral evidence. We additionally sample unrelated value constructs as \textit{unsupported} instances to strengthen the model's ability to reject irrelevant values. Together, these instances train the model to assess the evidential status of each construct rather than directly generate a label set. During inference, the model verifies each candidate value description against the behavior summary, and all constructs classified as \textit{supported} constitute the final multi-label prediction. This verification-based formulation enables B2V-Verifier to make conservative, evidence-aware predictions and reduces unsupported value attribution.

\section{Experiments}
\label{benchmark}
We evaluate the B2V task on our newly constructed benchmark, \textbf{B2V-Bench}. Given a user's pre-purchase behavioral trajectory and the corresponding item context, the task aims to predict the consumption value labels reflected in the behavioral episode. Formally, let \(E=\{a_1,a_2,\ldots,a_T\}\) denote a behavioral episode, where each action \(a_t\) contains the action type, item information, and contextual semantics. The goal is to predict a set of value labels \(Y \subseteq \mathcal{V}\), where \(\mathcal{V}\) denotes our E-commerce Consumption Value Taxonomy. Based on B2V-Bench, we benchmark traditional sequential models and mainstream LLMs and verify the effectiveness of our proposed B2V-Verifier for B2V measurement.


\vspace{-0.3em}
\subsection{Experimental setup}

\noindent\textbf{Models.} For sequential models, we train three representative architectures~\cite{hidasi2015session,zhou2018deep,sun2019bert4rec} on the original structured action sequences, replacing their item-prediction heads with \(K\)-way consumption-value classifiers. For LLMs, we benchmark 10 advanced models spanning GPT, Qwen, DeepSeek, Gemini, Claude, GLM, and Kimi. They use the behavioral summaries validated during annotation, which retain decision-relevant evidence while reducing redundant interactions, and are evaluated under four prompting settings \(P1\)--\(P4\) with increasing task guidance (see Appendix H). We train B2V-Verifier by supervised fine-tuning Qwen3-8B with VVT using LLaMA-Factory~\cite{zheng2024llamafactory} on four NVIDIA A40 GPUs (See Appendix I).



\noindent\textbf{Metrics.} We evaluate models from three perspectives: multi-label classification, label
ranking, and primary-label identification. Multi-label classification is
measured using Macro-F1 (\textit{Ma-F1}), Micro-F1 (\textit{Mi-F1}), Weighted-F1 (\textit{W-F1}), macro
Precision (\textit{Prec.}), macro Recall (\textit{Rec.}), and Hamming Loss
(\textit{Ham.})~\cite{sokolova2009systematic,schapire2000boostexter,tsoumakas2007multi}, assessing label-level correctness, coverage, and robustness under label imbalance.
To evaluate the quality of the ranking, we use NDCG@k ($k=1,2,3$), denoted N \textit{@ k}~\cite{jarvelin2002cumulated}. For primary-label identification, we report Primary Accuracy (\textit{Acc.}), Top-2 Accuracy
(\textit{T2 Acc.}), and their average.


\vspace{-0.3em}
\subsection{Main Results}

Sequential models show limited effectiveness on B2V-Bench. The poor performance of GRU4Rec, BERT4Rec, and DIN (See Appendix B) suggests that B2V requires more than modeling user--item interaction sequences alone. This is because the task requires reasoning over product attributes and behavioral actions. As shown in Table~\ref{tab:b2v_results}, LLMs substantially outperform sequential models, highlighting the importance of semantic understanding in behavior-to-value inference. However, adding more information to the prompt does not lead to continued performance gains. More detailed descriptions of value construct generally clarify their semantics and reduce over-attribution, whereas few-shot examples do not consistently improve performance. For some models, demonstrations bias predictions toward familiar label combinations, improving primary-label accuracy but reducing secondary-value recall. Among the LLM baselines, GPT-5 exhibits strong classification performance, while DeepSeek-V3.2 under P4 achieves the most balanced overall results. Nevertheless, all LLM baselines remain clearly behind B2V-Verifier, indicating that general-purpose semantic reasoning alone is insufficient for reliable B2V measurement.

Our B2V-Verifier achieves the best performance across all metrics. The large improvement in Macro-F1 is important, showing that the B2V-Verifier can better recognize diverse and less frequent value constructs rather than relying on dominant value labels. Its superior ranking and primary-label identification performance further demonstrate the effectiveness of VVT for more accurate B2V measurement.

\vspace{-0.3em}
\subsection{Ablation Study}

To assess the contribution of VVT's designs, we conduct an ablation study on B2V-Bench, comparing the full B2V-Verifier with four variants that remove the verification formulation, replace value descriptions with raw label names, or drop \textit{unsupported} and \textit{insufficient} modeling. All variants use the same training and evaluation settings. As shown in Table~\ref{tab:ablation}, the full VVT framework outperforms all ablations. Removing the verification formulation causes the largest drop, confirming the advantage of VVT over direct label prediction. The remaining drops further show that explicit value descriptions and fine-grained negative states help clarify construct boundaries and prevent unsupported value attribution.


\vspace{-0.3em}
\section{Conclusion}

In this work, we introduce the \textbf{Behavior-to-Value (B2V)} task, which aims to identify the consumer value tendencies manifested in an episode of e-commerce behavior. To support this task, we construct the \textbf{E-commerce Consumption Value Taxonomy} and introduce \textbf{B2V-Bench}, the first large-scale benchmark for measuring consumer values from e-commerce behaviors. We further present \textbf{B2V-Verifier}, a B2V measurement model based on Value Verification Tuning. Experiments show that it outperforms strong baselines, improving multi-label classification by 34\%. This work lays a foundation for value-aware user modeling in e-commerce and opens promising directions for future research in value-aware recommendation and interpretable user profiling.



\bibliography{custom}

\end{document}